\pdfoutput=1
\documentclass[11pt]{article}

\usepackage{lmodern}

\usepackage[preprint]{acl}

\usepackage{times}
\usepackage{latexsym}
\usepackage{amsmath}
\usepackage{amssymb}
\usepackage{booktabs}
\usepackage{dblfloatfix}
\usepackage{subcaption}
\usepackage{hyperref}
\usepackage{caption}

\usepackage{titlesec}

\usepackage[T1]{fontenc}

\usepackage[utf8]{inputenc}

\usepackage{microtype}

\usepackage{inconsolata}

\usepackage{graphicx}

\title{Beyond Local Surprise: Grounded Dialogue as Selective Belief Revision under Referential Uncertainty
}

\author{
\textbf{Ziming Liu\textsuperscript{1}},
\textbf{Bhanu Chaitanya Jasti\textsuperscript{1}},
\textbf{Ziyang Xu\textsuperscript{2}},
\textbf{Hongyu Wu\textsuperscript{1}},
\textbf{Yi Wu\textsuperscript{1}},
\textbf{Jiqun Liu\textsuperscript{3}}
\\
\\
\textsuperscript{1}School of Computer Science, University of Oklahoma \\
\textsuperscript{2}School of Library and Information Studies, University of Oklahoma
\\
\textsuperscript{3}School of Information Studies, University of Wisconsin–Milwaukee\\
\small{
\textbf{Correspondence:} \href{mailto:ziming.liu-1@ou.edu}{ziming.liu-1@ou.edu}
}
}

\begin{document}
\maketitle
\begin{abstract}
\setlength{\spaceskip}{0.92\fontdimen2\font plus 0.5\fontdimen3\font minus 0.5\fontdimen4\font}

\titlespacing{\section}
{0pt}{1.2ex plus .2ex minus .2ex}{0.6ex}

\titlespacing{\subsection}
{0pt}{0.9ex plus .2ex minus .2ex}{0.4ex}

When a speaker refers to a scene that the listener cannot directly see, the listener must decide whether to preserve its current understanding or revise it as new utterances arrive. Many language systems treat local mismatch as a cue for updating: divergence from the current understanding encourages adjustment. Yet conversational understanding may be more conservative, interpreting mismatching evidence relative to prior understanding rather than immediately revising it. We introduce a controlled, data-driven framework for turn-by-turn preserve/revise decisions in dialogue, where competing revision policies are learned under otherwise identical conditions. We compare four theory-driven revision strategies, each reflecting a different assumption about when listeners should preserve or revise. Two findings stand out. First, a mismatch-driven policy that updates solely based on local divergence reacts strongly to mismatch but destabilizes grounding and degrades retrieval. Second, an uncertainty-sensitive policy extends mismatch-based updating with accumulated evidence, preserving coherent understanding while maintaining strong retrieval performance. Surprisingly, coherent understanding emerges from a counterintuitive pattern: local mismatch promotes preservation, whereas accumulated uncertainty promotes revision, suggesting that listeners maintain prior understanding despite local mismatch and revise only when uncertainty sufficiently accumulates. This pattern is consistent with conceptual pact theory.

\end{abstract}

\section{Introduction}


Human communication succeeds because listeners must continually decide whether to preserve an established interpretation or revise it as new conversational evidence arrives. For example, when a speaker mentions \textit{my new pet}, a listener may initially picture a dog; when a later turn references \textit{her tank}, that interpretation may shift toward a fish or reptile. This process, known as conversational grounding \cite{clark1989contributing, clark1991grounding}, is inherently incremental: each conversational turn is interpreted relative to what has already been established between speakers~\cite{shaikh2024grounding}. Grounding therefore depends not only on updating understanding, but on deciding when prior interpretations should be preserved and when they should be revised. By constructing systems that solve this grounding problem turn by turn, we can use their behavior to study how interpretations are preserved and revised over time.

\begin{figure*}[t]
    \centering
    \includegraphics[width=\textwidth]{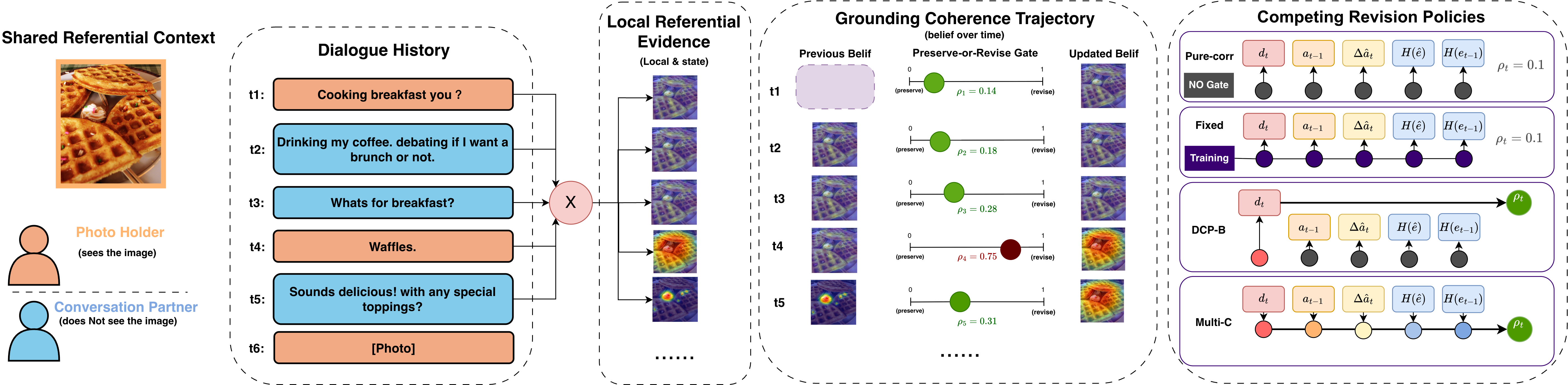}
    \caption{
    Overview of the preserve-or-revise framework for grounded dialogue. 
    Grounding evolves over dialogue turns by integrating local referential evidence with prior belief through a preserve-or-revise mechanism. 
    The underlying architecture is fixed while revision policies vary: \textsc{pure-corr}, \textsc{fixed}, \textsc{dcp-b}, and \textsc{multi-c}.
    }
    \label{fig:framework}
\end{figure*}

Many computational approaches to grounded language understanding model interpretation as an incremental update process, where interpretations are continuously revised as new evidence arrives. In incremental reference resolution, for example, models update a probability distribution over candidate referents as each new word narrows or shifts the interpretation \cite{kennington2017simple}. Consider the referring expression \textit{red book}: hearing \textit{red} initially activates a broad set of possible objects, while \textit{book} subsequently narrows the listener’s belief toward a more specific referent \cite{yule2013referential}. Implicit in many such approaches is a simple computational intuition: larger mismatch should warrant stronger revision of current understanding. Yet whether mismatch alone should determine when and how much listeners revise an established interpretation remains unclear. Incremental models demonstrate that understanding evolves over time, but leave open the question of how revision should be driven. 

Conceptual pact theory suggests that conversational partners preserve shared interpretations rather than renegotiating meaning at every local mismatch \cite{brennan1996conceptual}. This raises a different computational possibility: mismatch alone may not be sufficient to trigger revision. Instead, effective grounding may depend not on isolated mismatch, but on whether repeated mismatch accumulates into uncertainty about current understanding. To investigate this idea, we introduce a simulation framework that compares competing revision policies in grounded visual dialogue (Figure~\ref{fig:framework}) \cite{zang2021photochat}. Across revision policies, a clear pattern emerges: reacting strongly to local mismatch destabilizes grounding. Although local divergence reliably detects mismatch, using it as the primary trigger for revision repeatedly disrupts accumulated understanding. In contrast, stable grounding emerges when revision is guided by uncertainty over accumulated evidence. Unexpectedly, uncertainty promotes revision while surprise suppresses it, suggesting that listeners preserve interpretations despite local mismatch and revise only when uncertainty sufficiently accumulates. 

\section{Related Work}
Conversational grounding is commonly understood as an incremental process through which interlocutors establish shared understanding over time \cite{clark1989contributing,clark1991grounding}. When misunderstandings arise, prior work often frames grounding in terms of conversational repair, emphasizing clarification and recovery following confusion or communication breakdown \cite{schegloff1977preference, traum1994computational}. Dialogue systems similarly treat revision as a discrete decision about whether clarification is required under uncertainty \cite{shi2022learning,testoni2024asking}. Our work differs in focusing not on whether repair occurs, but on the computational assumptions governing when grounding should be preserved or revised.

A complementary line of work studies how listeners revise interpretations incrementally as linguistic evidence accumulates \cite{tanenhaus1995integration, altmann1999incremental, kennington2017simple}. In situated reference resolution and grounded dialogue, models update interpretations incrementally as new evidence arrives, gradually narrowing or shifting toward likely referents \cite{kennington2017simple, schlangen2011general, li2016reference}. A common intuition in these approaches is simple: larger mismatch should lead to larger updates~\cite{meister2024towards}. However, prior work typically assumes a revision mechanism rather than treating revision itself as the object of study. While incremental grounding and repair have been extensively modeled, preserve-or-revise behavior is rarely formulated as an explicit computational question. We instead directly compare competing assumptions about how revision should be determined under mismatch and uncertainty.

Grounded multimodal dialogue provides a natural setting for studying conversational grounding because interlocutors must align language against a shared visual environment over time \cite{das2017visual,kim2019codraw,zang2021photochat}. Prior work typically evaluates grounding through downstream task success such as retrieval or collaborative performance \cite{zang2021photochat,willemsen2023resolving,lee2024stark,fried2023pragmatics}. Rather than treating grounded dialogue only as a task, we leverage it as a controlled setting for demonstrating how different revision assumptions shape grounding under identical interactions. This allows grounding behavior to be compared independently of changes in perceptual representations or interaction context.

\section{Method}
Figure~\ref{fig:framework} provides an overview of the preserve-or-revise framework. Grounding evolves through turn-level evidence accumulation over a shared visual scene, while competing revision policies differ only in how preserve-or-revise decisions are determined.

\paragraph{Grounding as an Evolving Evidence State.}
We model conversational grounding as an evolving evidence state over image regions, allowing preserve-or-revise behavior to be analyzed turn by turn. The shared image provides a stable referential anchor, creating a controlled setting in which different revision assumptions can be examined under identical interactions.

Given a dialogue--image pair $(D, I)$, where $D={u_t}_{t=1}^{T}$ denotes a sequence of utterances and $I$ a shared visual scene, the goal is to track how grounding evolves over time rather than simply determine whether dialogue and image match. At each turn $t$, the model maintains an evidence state $e_t$ over image regions, representing its current estimate of what has been grounded.

At each turn, the model determines how strongly new linguistic evidence should update the accumulated grounding state. We model this through a scalar revision signal $\rho_t \in (0,1)$ that balances prior grounding against newly observed evidence: low $\rho_t$ preserves accumulated grounding, whereas high $\rho_t$ favors revision toward new evidence. By varying how $\rho_t$ is determined while holding dialogue and visual evidence fixed, the framework isolates competing assumptions about when revision should occur.

\paragraph{Perception as a Fixed Prior.} To isolate grounding from perceptual learning, we keep a frozen CLIP encoder (ViT-B/32; \cite{radford2021learning}) fixed throughout training. This means the model’s visual–linguistic knowledge remains unchanged, while only grounding is allowed to evolve over dialogue. Within conversation, the image serves as a stable referential anchor that new language is grounded against turn by turn. As new utterances arrive, grounding updates over image regions while visual representations remain fixed, allowing different revision policies to be examined under identical perceptual conditions. Utterances are encoded using CLIP text features and images using global and patch-level visual features. A shared trainable projection $W_h$ maps language into the visual space for grounding attention, coherence estimation, and contrastive alignment, while visual representations are reused from frozen CLIP.

\paragraph{Turn-local evidence and accumulated grounding state.} Each utterance provides new evidence about which parts of the shared scene are currently referenced. We compute this turn-local signal as a softmax attention distribution over patch-level image features:
\begin{equation}
\hat{e}_t = \mathrm{softmax}(q_t^\top V)
\end{equation}
where $\hat{e}_t \in \Delta^{49}$ captures where the current utterance points within the visual scene. This signal is intentionally local: it reflects only what the current utterance demands, without knowledge of prior conversational context.

Grounding is not determined by a single utterance alone. The accumulated grounding state $e_t$ integrates turn-local evidence with prior grounding through a revision-weighted interpolation:
\begin{equation}
e_t = (1-\rho_t)\,e_{t-1} + \rho_t\,\hat{e}_t
\end{equation}
initialized from a uniform prior $e_0 = \frac{1}{49}\mathbf{1}$. The scalar $\rho_t \in (0,1)$ controls how strongly new evidence revises grounding at each turn. Different assumptions about $\rho_t$ therefore capture different ways listeners may preserve or revise over time.

\textbf{Coherence as an observable grounding trajectory.}
The evidence state $e_t$ captures what is currently grounded, but does not by itself indicate whether grounding remains coherent across turns. We therefore track grounding coherence as an evolving trajectory. An evidence-weighted visual summary $c_t = e_t^\top V$ combines the current grounding state with image features, and turn-level coherence is estimated as:
\begin{equation}
\tilde{a}_t = \sigma\!\left(W_a[q_t;\,c_t;\,v^{\mathrm{cls}}]
\right)
\end{equation}
We then update coherence over time:
\begin{equation}
a_t = (1-\lambda_t)\,a_{t-1} + \lambda_t\,\tilde{a}_t, 
\quad \lambda_t = 1 - \rho_t
\end{equation}
This coupling allows local evidence and longer-term coherence to be tracked separately. The resulting signal $a_t$ provides an interpretable trajectory of grounding coherence across dialogue turns.

\textbf{Different policies about grounding revision.} Rather than assuming a single mechanism for preserve-or-revise decisions, we compare four competing policies about how listeners should revise grounding over time. Each hypothesis represents a different assumption about the role of mismatch, uncertainty, and prior grounding stability under identical interaction settings. Comparing their behavior allows us to measure not only whether grounding succeeds, but what kinds of revision dynamics support coherent grounding.

\begin{itemize}

\item \textsc{pure-corr}: \textbf{Grounding as conservative accumulation.}
Grounding evolves through weak, fixed accumulation ($\rho=0.1$) without explicit preserve-or-revise supervision. This policy assumes that understanding changes gradually and serves as a minimal baseline for grounding dynamics.

\item \textsc{fixed}: \textbf{Grounding as constant-rate revision.}
Listeners revise grounding at a fixed pace regardless of conversational context, implemented as a constant revision rate ($\rho = 0.1$) across all turns. Unlike \textsc{pure-corr}, preserve-or-revise objectives remain active, allowing grounding coherence to shape learning while revision strength itself stays fixed. This reflects the assumption that grounding should evolve steadily rather than adaptively.

\item \textsc{dcp-b}: \textbf{Grounding as surprise-driven revision.}
Revision is governed by mismatch between prior grounding and incoming evidence:
\begin{equation}
d_t = \mathrm{KL}(e_{t-1} \| \hat{e}_t)
\end{equation}
Larger divergence increases the revision signal $\rho_t$, shifting grounding toward new evidence. This reflects a common computational intuition: local mismatch should trigger stronger updating.

\item \textsc{multi-c}: \textbf{Grounding as uncertainty-sensitive revision.}
Rather than treating local mismatch as sufficient for revision, this policy assumes that preserve-or-revise decisions depend on broader contextual signals: prior grounding stability, uncertainty in current understanding, and whether trusting new evidence would improve coherence. Revision is therefore modeled as a learned function of five conversational signals:
\begin{equation}
\scalebox{0.82}{$\displaystyle
\rho_t = \sigma\!\left(
W_p\left[
d_t,\,
a_{t-1},\,
\Delta\hat{a}_t,\,
H(e_{t-1}),\,
H(\hat{e}_t)
\right]
\right)$}
\end{equation}
where $d_t$ measures local evidence divergence, $a_{t-1}$ prior grounding coherence, $\Delta\hat{a}_t$ the expected coherence change if new evidence were trusted, and $H(\cdot)$ uncertainty in prior and incoming grounding. To avoid circular dependence, $\Delta\hat{a}_t$ is estimated from a counterfactual fully-trusted grounding state before updating occurs:
\begin{equation}
\scalebox{0.82}{$
\hat{a}_t
=
\sigma\!\left(
W_a[q_t;\,\hat{c}_t;\,v^{\mathrm{cls}}]
\right),
\qquad
\Delta\hat{a}_t
=
\hat{a}_t-a_{t-1}
$}
\end{equation}
This parameterization evaluates whether revision should depend on multiple contextual signals rather than local divergence alone.

\end{itemize}

\paragraph{Learning grounding dynamics.}
Training is guided by preserve-or-revise constraints rather than explicit repair supervision. The objectives encourage stable grounding during low-revision turns and measurable grounding change during high-revision turns, while keeping perceptual evidence fixed.

\begin{itemize}

\item \textbf{Grounding coherence through contrastive alignment.}
To ensure that grounding remains anchored to the shared visual scene over time, we optimize a cross-dialogue contrastive objective. At each turn, an utterance query $q_t$ is encouraged to remain aligned with its corresponding dialogue image while staying distinguishable from images in other dialogues:
\begin{equation} \small
L_{\mathrm{corr}} =
-\sum_t \log
\frac{\exp(s_t^+/\tau)}
{\sum_b \exp(s_b/\tau)}
\end{equation}
where
$s_t^+ = \mathrm{sim}(q_t, v^{\mathrm{cls}}_{d(t)})$
demonstrates similarity to the correct image,
$s_b = \mathrm{sim}(q_t, v_b^{\mathrm{cls}})$
to negatives, and $\tau$ is a temperature parameter. This objective stabilizes grounding against the shared visual context, ensuring that preserve-or-revise dynamics unfold relative to a consistent referential target.

\item \textbf{Preserve when evidence remains consistent.}
The preserve objective penalizes unnecessary shifts during low-revision turns. When revision pressure is low, accumulated grounding is encouraged to remain stable rather than drift unnecessarily. We operationalize this through a preserve constraint that penalizes changes in grounding during low-revision turns:
\begin{equation}
\scalebox{0.75}{$
L_{\mathrm{preserve}}
=
\frac{1}{T-1}
\sum_{t\ge2}
(1-\rho_t)_{\mathrm{detach}}
\,
\mathrm{KL}(e_{t-1}\|e_t)
$}
\end{equation}
Here, $\rho_t$ is detached so that revision acts as a state regulator rather than a directly optimized objective.

\item \textbf{Revise when accumulated evidence warrants change.}
The revise objective encourages grounding to change when revision pressure is high. When revision pressure is high, grounding is encouraged to change rather than remain artificially stable. We measure this through a revise constraint that penalizes turns where accumulated grounding changes too little despite strong revision signals:
\begin{equation}
\scalebox{0.65}{$
L_{\mathrm{revise}}
=
\frac{1}{T-1}
\sum_{t\ge2}
\rho_{t,\mathrm{detach}}
\,
\max\!\Big(
0,\,
\delta
-
\mathrm{KL}(e_{t-1}\|e_t)
\Big)
$}
\end{equation}
where $\delta$ defines the minimum grounding shift expected when revision pressure is high. As in the preserve objective, $\rho_t$ is detached so that revision remains a state-dependent regulator rather than a trivially optimized target.

\item \textbf{Selective revision rather than constant updating.}
 We regularize the mean revision rate toward a target frequency $r$ to encourage sparse revision. We therefore regularize the mean revision rate toward a target frequency $r$, encouraging revision to remain sparse rather than constant:
\begin{equation}
L_{\mathrm{rate}}
=
\big(\mathrm{mean}_t(\rho_t)-r\big)^2
\end{equation}
This objective operationalizes a preserve-by-default assumption, in which revision remains available but is expected to occur selectively rather than continuously.

\end{itemize}

The full objective combines these terms:
\begin{equation} \small
\mathcal{L} = L_{\mathrm{corr}} + \alpha L_{\mathrm{preserve}}
+ \beta L_{\mathrm{revise}} + \gamma L_{\mathrm{rate}}
\end{equation}
encouraging stable grounding under consistent evidence with selective revision when needed.

\section{Experimental Setup}

\paragraph{Task and Dataset.} We evaluate conversational grounding on PhotoChat \citep{zang2021photochat}, a grounded dialogue dataset in which one participant privately sees an image and reveals it only after a conversation unfolds. This creates referential uncertainty: interlocutors must establish shared understanding through dialogue alone, without direct access to the image. Because reference is refined across turns, successful interaction depends on coherent grounding over time rather than isolated utterance matching, making PhotoChat a natural setting for studying how grounding is preserved and revised under referential uncertainty.

To isolate incremental grounding, we retain only turns before image disclosure and treat the hidden image as the grounding target. This prevents trivial matching from explicit visual descriptions and requires grounding to evolve through accumulated conversational evidence across turns. The dataset contains 21 training shards. We use a deterministic shard-level split (19 train, 2 test; random seed 42), fixed across all experiments to ensure comparability across revision policies. Full hyperparameter settings are provided in Appendix~\ref{sec:appendix}.

\begin{table*}[h]
\centering
\scriptsize
\setlength{\tabcolsep}{4pt}
\begin{tabular}{lcccccccc}
\toprule
& \multicolumn{5}{c}{\textit{Retrieval}} 
& \multicolumn{3}{c}{\textit{Grounding Dynamics}} \\
\cmidrule(lr){2-6}\cmidrule(lr){7-9}
\textbf{Regime}
& \textbf{R@1}
& \textbf{R@5}
& \textbf{R@10}
& \textbf{MRR}
& \textbf{MedR}$\downarrow$
& \textbf{Pres.}$\downarrow$
& \textbf{Rev. Sens.}$\uparrow$
& \textbf{Mean $\rho$} \\
\midrule
Random
& 0.0011 & 0.0056 & 0.0112 & 0.0083 & 447
& --- & --- & --- \\
\textsc{clip-only}
& 0.0873 & 0.2066 & 0.2630 & 0.1450 & 73
& --- & --- & --- \\
\midrule
\textsc{pure-corr}
& 0.118 $\pm$ .005 & 0.314 $\pm$ .005 & 0.424 $\pm$ .006
& 0.219 $\pm$ .006 & 16.0 $\pm$ 1.0
& 0.050 $\pm$ .000 & --- & 0.100 \\
\textsc{fixed}
& 0.117 $\pm$ .006 & 0.315 $\pm$ .003 & 0.426 $\pm$ .006
& 0.218 $\pm$ .007 & 16.0 $\pm$ 1.0
& 0.049 $\pm$ .001 & --- & 0.100 \\
\textsc{dcp-b}
& 0.009 $\pm$ .006 & 0.052 $\pm$ .032 & 0.086 $\pm$ .048
& 0.038 $\pm$ .019 & 184.0 $\pm$ 121.3
& 0.001 $\pm$ .001 & 0.900 $\pm$ .015 & 0.203 \\
\textsc{multi-c}
& 0.112 $\pm$ .004 & 0.304 $\pm$ .012 & 0.418 $\pm$ .009
& 0.213 $\pm$ .003 & 16.0 $\pm$ 0.0
& 0.083 $\pm$ .001 & 0.598 $\pm$ .097 & 0.148 \\
\bottomrule
\end{tabular}
\caption{
Retrieval performance and intrinsic grounding dynamics across policies (mean $\pm$ std, three seeds). \textsc{pure-corr} and \textsc{fixed} remain within one standard deviation of \textsc{multi-c} on all retrieval metrics. \textsc{dcp-b} exhibits the strongest intrinsic dynamics, including large mismatch-triggered switches (Switch KL $=12.31 \pm 0.88$), yet substantially underperforms on retrieval.
}
\label{tab:retrieval}
\end{table*}

\section{Evaluation Protocol}
We evaluate grounding quality through visual retrieval. Given a dialogue trajectory, the model retrieves the target image from a gallery of candidates. Retrieval serves as a behavioral test of grounding quality rather than grounding itself: a model may retrieve the correct image, yet rely on unstable or incoherent revision dynamics. At the same time, grounding that fails to remain coherent should eventually impair retrieval. We therefore evaluate both behavioral performance and grounding dynamics, asking not only whether a model succeeds, but how that success is achieved. We report Recall@$K$ ($R@1$, $R@5$, $R@10$), median rank (MedR), and mean reciprocal rank (MRR), together with revision behavior and coherence trajectories.

\paragraph{Reference Conditions.}
We include two reference conditions to contextualize grounding behavior under the four revision policies (\textsc{pure-corr}, \textsc{fixed}, \textsc{dcp-b}, \textsc{multi-c}):

\begin{itemize}

\item \textbf{Random}: chance-level retrieval without grounding, providing a lower-bound.

\item \textbf{CLIP-only}: frozen CLIP representations without grounding updates or task-specific training. Utterances are encoded independently and mean-pooled into a dialogue-level query, then matched to frozen image embeddings via cosine similarity. This provides a strong static reference without incremental grounding dynamics.

\end{itemize}

These conditions serve as behavioral anchors for interpreting how different revision policies shape grounding and retrieval.

\subsection{Intrinsic Grounding Dynamics}
To characterize how grounding evolves over time, we analyze three intrinsic metrics capturing complementary aspects of preserve-or-revise behavior. All metrics are computed over evaluation turns and aggregated across random seeds.

\paragraph{Preservation index (Pres).}
We measure grounding stability during low-revision turns ($\rho_t < 0.5$) as the mean evidence shift between consecutive grounding states:
$\mathrm{KL}(e_{t-1}\|e_t)$.
Lower values indicate stronger preservation of prior grounding when revision pressure is low.

\paragraph{Revision sensitivity (Rev Sens).}
We measure how strongly revision signals correspond to realized grounding updates by computing the Pearson correlation between $\rho_t$ and the evidence shift $\mathrm{KL}(e_{t-1}\|e_t)$. Higher values indicate that larger revision signals are associated with larger grounding changes. This metric is undefined for \textsc{pure-corr} and \textsc{fixed}, which use constant $\rho_t$.

\paragraph{Mean revision rate (Mean $\rho$).}
We report the mean revision rate $\rho_t$ across turns, characterizing whether grounding evolves selectively or continuously. For \textsc{dcp-b}, we additionally report the average grounding shift (Switch KL) during high-revision turns ($\rho_t \geq 0.5$), capturing how strongly grounding changes once mismatch-driven revision is activated.

\captionsetup{
    font=small,
    skip=3pt
}

\begin{figure*}[t]
\centering

\begin{minipage}[t]{0.64\textwidth}
    \centering

    \begin{minipage}[t]{0.495\linewidth}
        \centering
        \includegraphics[width=0.8\linewidth]{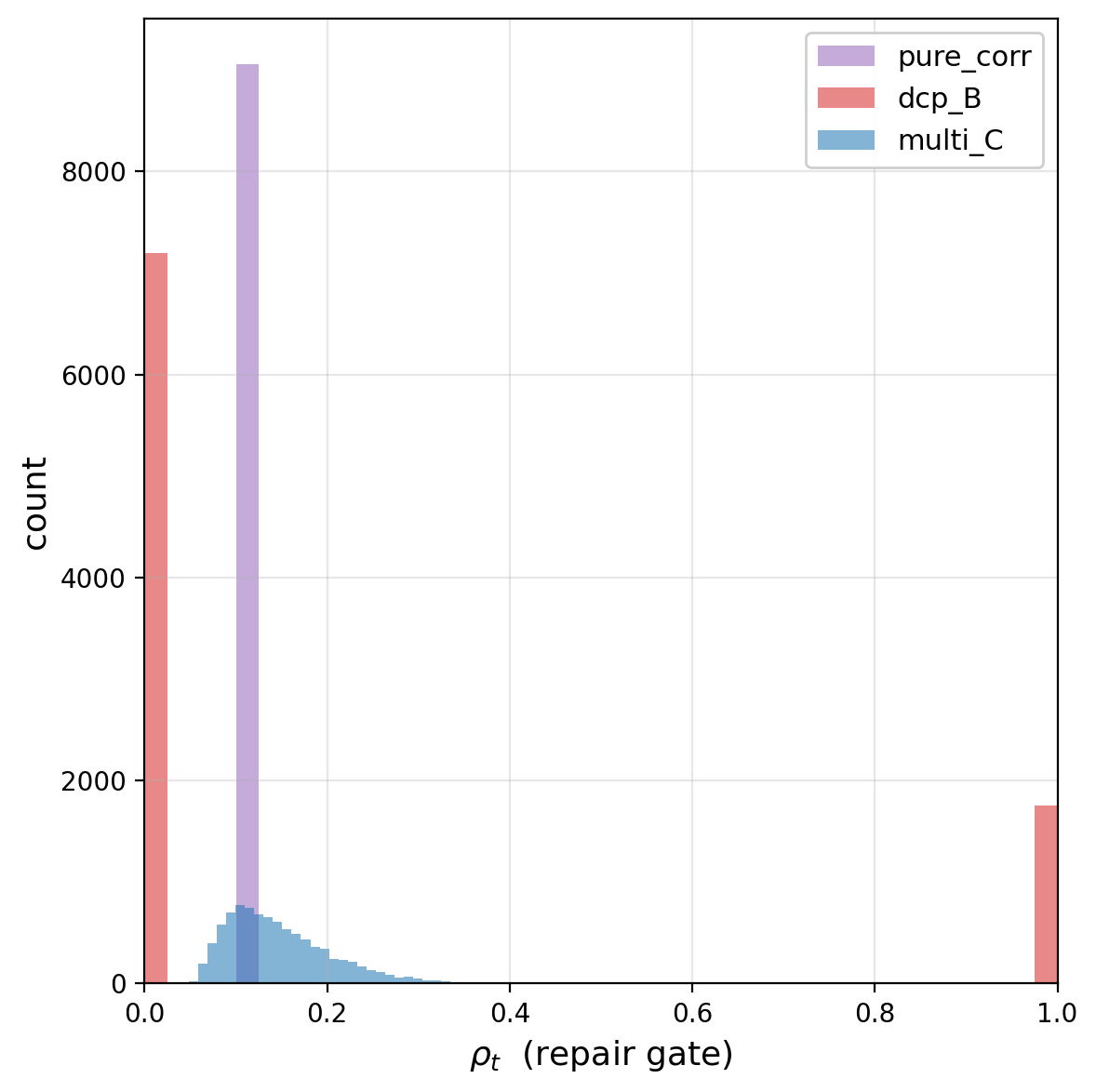}
    \end{minipage}
    \hfill
    \begin{minipage}[t]{0.495\linewidth}
        \centering
        \includegraphics[width=0.8\linewidth]{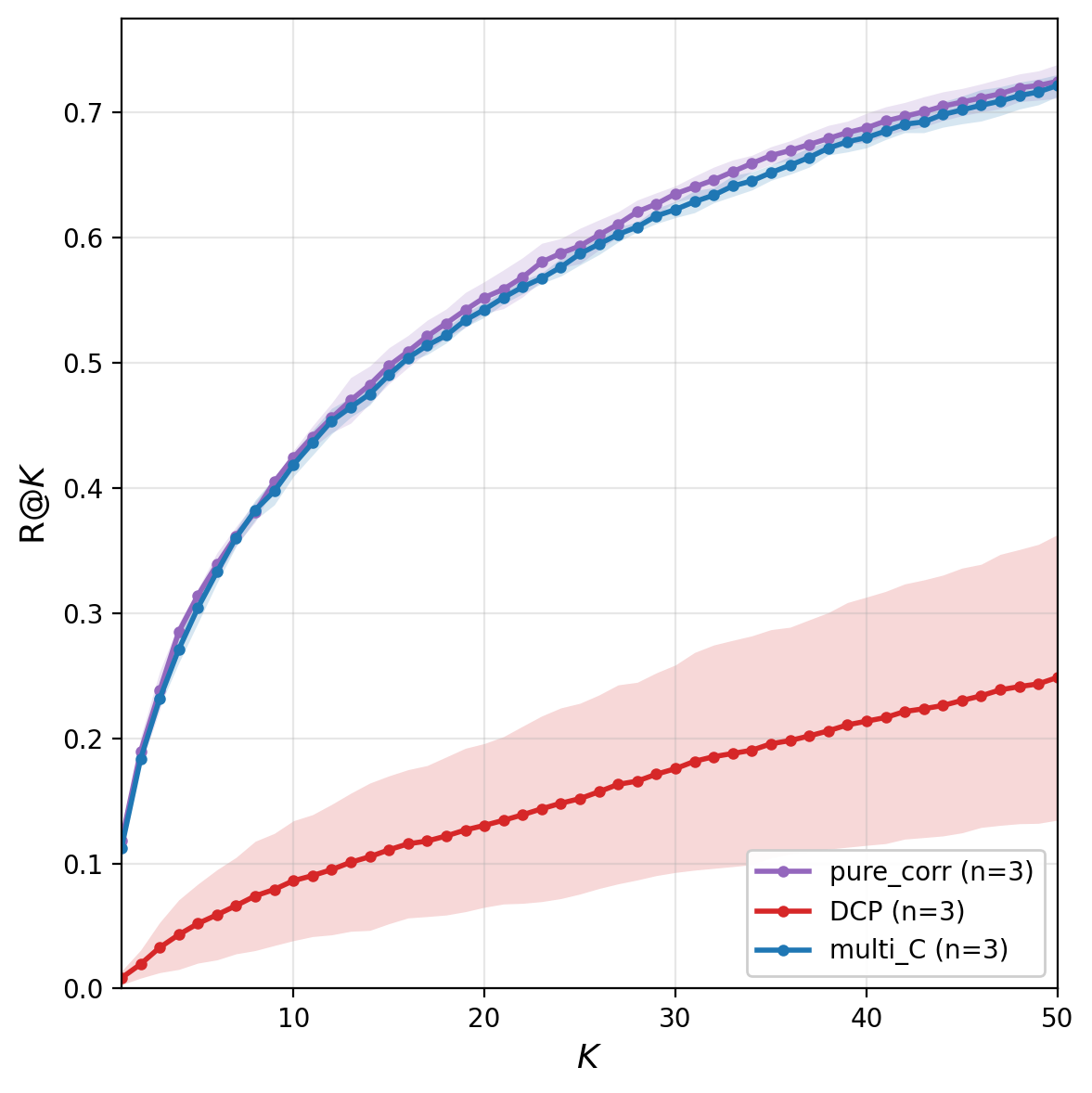}
    \end{minipage}

    \vspace{-0.35em}

    \caption{
    \textit{Left}: Learned revision policies. 
    \textsc{pure-corr} remains low and deterministic, 
    \textsc{dcp-b} collapses into near-binary switching, 
    and \textsc{multi-c} maintains bounded revision. 
    \textit{Right}: R@$K$ retrieval performance. 
    \textsc{multi-c} closely matches \textsc{pure-corr}, 
    while \textsc{dcp-b} consistently underperforms.
    }
    \label{fig:dynamics_results}
\end{minipage}
\hfill
\begin{minipage}[t]{0.32\textwidth}
    \vspace{-6.5 em}
    \centering
    \scriptsize
    \setlength{\tabcolsep}{3pt}
    \renewcommand{\arraystretch}{0.95}

    \begin{tabular}{lcc}
    \toprule
    \textbf{Signal}
    & \textbf{Contrib.}
    & \textbf{Corr} \\
    \midrule
    $d_t$             & 0.399 & $-$0.935 \\
    $a_{t-1}$         & 0.034 & $+$0.040 \\
    $\Delta\hat{a}_t$ & 0.025 & $-$0.008 \\
    $H(e_{t-1})$      & 0.054 & $-$0.176 \\
    $H(\hat{e}_t)$    & 0.103 & $+$0.562 \\
    \bottomrule
    \end{tabular}

    \vspace{1.4 em}

    \captionof{table}{
    Signal contributions to \textsc{multi-c} revision and correlation with $\rho_t$ (mean across seeds). Contribution is computed as scaled importance, $|W_i|\cdot \mathrm{std}(x_i)$.
    }
    \label{tab:feature_contrib}
\end{minipage}

\vspace{-0.6em}
\end{figure*}

\subsection{Modality Dependence} To verify that learned grounding dynamics reflect actual multimodal coordination rather than single-modality shortcuts, we perform inference-time perturbation experiments on \textsc{multi-c}. We selectively replace text or image representations with zero vectors, random Gaussian vectors, or empirical mean embeddings without retraining. If grounding emerges from coordinated preserve-or-revise behavior across modalities, perturbing either modality should impair retrieval toward chance. Differences across perturbation types further reveal how strongly grounding depends on linguistic versus visual evidence under different conditions.

\section{Results}

\paragraph{Retrieval Performance and Grounding Dynamics.}
Table~\ref{tab:retrieval} reveals an important contrast. Despite different grounding dynamics, \textsc{pure-corr}, \textsc{fixed}, and \textsc{multi-c} achieve comparable retrieval performance across metrics. This suggests that successful grounding does not require aggressive revision, as long as accumulated understanding remains coherent over time.

\textsc{dcp-b}, however, collapses on retrieval. Although its revision signal closely tracks local evidence change (Rev. Sens. $=0.900$), retrieval performance deteriorates substantially. Its intrinsic dynamics reveal why: once local mismatch triggers revision, grounding shifts abruptly (Switch KL $=12.31 \pm 0.88$), repeatedly disrupting accumulated understanding. Rather than supporting coherent grounding, local mismatch drives overly reactive belief updates. High sensitivity to conversational change alone is therefore insufficient for successful grounding.

\textsc{multi-c} resolves this tradeoff. Retrieval remains within standard deviation of \textsc{pure-corr} and \textsc{fixed}, while maintaining structured revision behavior (Rev. Sens. $=0.598$, Mean $\rho=0.148$). Rather than reacting strongly to isolated mismatch, revision remains selective and distributed across turns, preserving accumulated grounding while remaining responsive to new evidence.

\paragraph{Local divergence is associated with reduced revision.}
Table~\ref{tab:feature_contrib} reveals a counterintuitive pattern. Contrary to the common intuition that local mismatch should trigger stronger updating, local divergence $d_t$ correlates negatively with revision in \textsc{multi-c} ($r=-0.935$), despite contributing strongly to the model (0.399). Higher divergence is therefore associated with lower revision rates rather than stronger updating. In contrast, incoming uncertainty $H(\hat{e}_t)$ shows the strongest positive association with revision ($r=+0.562$, contribution 0.103), suggesting that revision is influenced by ambiguity in new evidence rather than local mismatch alone. More broadly, the learned controller distributes revision across multiple conversational signals rather than relying primarily on divergence.

\paragraph{Different policies produce distinct revision behavior.}
Figure~\ref{fig:dynamics_results} (Left) shows that different competing policies produce distinct revision behavior. By design, \textsc{pure-corr} maintains fixed low revision, \textsc{dcp-b} collapses into near-binary switching driven by local mismatch, and \textsc{multi-c} learns smooth bounded revision across turns. These differences reveal fundamentally different revision assumptions: stable accumulation (\textsc{pure-corr}), surprise-driven switching (\textsc{dcp-b}), and uncertainty-sensitive revision (\textsc{multi-c}).

\begin{figure}[t]
\centering

\includegraphics[width=0.95\linewidth]{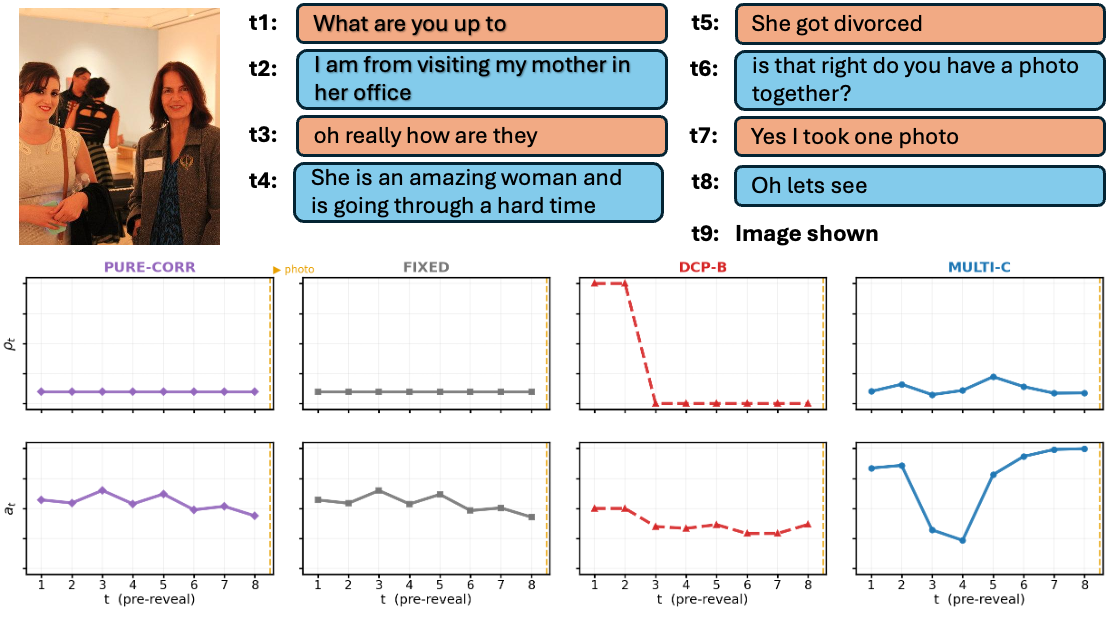}

\vspace{-0.4em}

\includegraphics[width=0.95\linewidth]{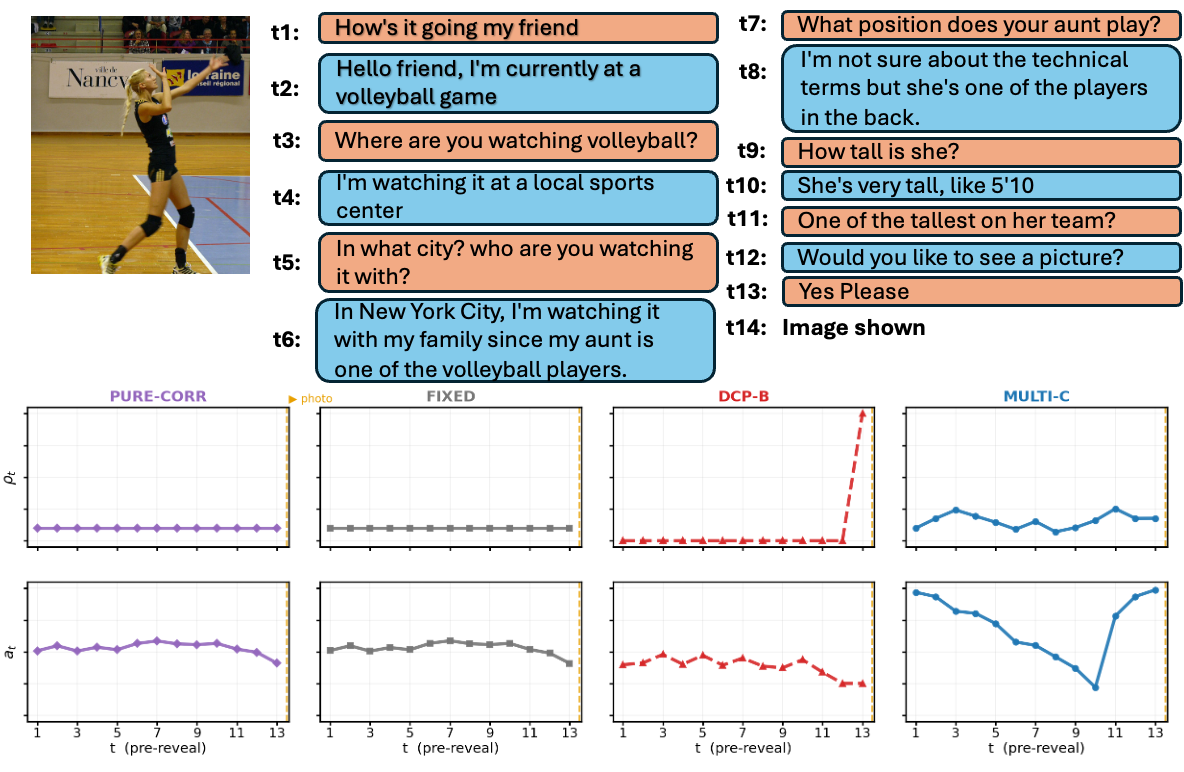}

\caption{
Turn-level grounding trajectories in two pre-reveal dialogues. \textsc{pure-corr} and \textsc{fixed} remain stable, \textsc{dcp-b} exhibits spike--freeze revision, and \textsc{multi-c} shows bounded adaptation to conversational change.
}
\label{fig:case_studies}

\vspace{-0.8em}
\end{figure}

\paragraph{Trajectories confirm the dissociation.}
Figure~\ref{fig:case_studies} presents two randomly sampled dialogues truncated before image reveal. The first involves a personal disclosure (\textit{``t5: She got divorced''}); the second establishes a volleyball-related reference early and revisits it through later conversational details before photo sharing.

In both dialogues, grounding assumptions produce distinct behaviors. By design, \textsc{pure-corr} and \textsc{fixed} maintain nearly constant revision and flat coherence trajectories, reflecting stable accumulation with limited adaptation. \textsc{dcp-b} instead exhibits a spike-then-freeze pattern: $\rho_t$ rises sharply at the first divergent turn and quickly collapses toward 0. Coherence drops and fails to recover despite continued evidence, suggesting that local mismatch triggers abrupt revision followed by disengagement from grounding.

In contrast, \textsc{multi-c} remains responsive without collapsing. In both dialogues, coherence temporarily decreases at conversational turning points but recovers as additional evidence accumulates, producing trajectories aligned with the evolving dialogue. Rather than collapsing after local mismatch, revision remains active but bounded, preserving accumulated grounding while adapting to new information. These trajectories provide an interpretable view of grounding pressure and recovery that is not visible from retrieval metrics alone.

\paragraph{Both modalities are necessary.}Inference-time perturbation experiments confirm that \textsc{multi-c} relies on both visual and linguistic information. Replacing either modality reduces retrieval to near-random levels across metrics (Appendix~\ref{sec:appendixb}). Grounding dynamics therefore cannot emerge from either modality alone.

\section{Discussion}

\paragraph{Successful grounding is conservative by default.}
\textsc{multi-c} converges to an assumption in which accumulation dominates and revision remains selective (Mean $\rho_t = 0.148$). Contrary to the common computational intuition that larger mismatch should trigger stronger updating, local divergence tends to reduce revision rather than amplify it. Revision instead appears to be more strongly associated with uncertainty in incoming evidence than with local conflict alone. This behavior resembles how interlocutors maintain common ground, preserving interpretations unless ambiguity becomes sufficiently high to motivate reinterpretation. Notably, the model recovers this pattern without explicit supervision. The result is broadly consistent with psycholinguistic accounts of conceptual pacts \cite{brennan1996conceptual}, in which partner-specific interpretations are preserved by default rather than continuously renegotiated.

\paragraph{Repair is not reducible to local mismatch.}
\textsc{dcp-b} suggests that sensitivity to conversational change alone may be insufficient for successful grounding. Although the model closely tracks local divergence (Rev Sens. = 0.900), grounding becomes unstable once mismatch triggers abrupt revision. This pattern suggests a limitation of relying on local divergence alone as a signal: mismatch does not always warrant reinterpretation. Divergence may instead reflect ambiguity, incomplete reference, or temporary conversational repair that still benefits from preserving prior grounding. Detecting mismatch and deciding whether mismatch justifies revision therefore appear to be distinct operations.

\paragraph{Grounding dynamics are not visible in retrieval alone.}
\textsc{multi-c} and \textsc{pure-corr} achieve comparable retrieval performance while exhibiting qualitatively different revision behavior. Retrieval indicates whether grounding succeeds, but not how it succeeds. Compared with \textsc{pure-corr}, \textsc{multi-c} exhibits selective preserve-or-revise behavior despite comparable retrieval, suggesting that similar behavioral outcomes can emerge from different grounding mechanisms. The case studies further suggest that these dynamics are behaviorally meaningful: coherence trajectories temporarily destabilize at conversational shifts and recover as evidence accumulates. The mechanisms separating successful grounding from failure therefore provide additional evidence about how conversational grounding unfolds over time. While these findings are consistent with uncertainty-sensitive belief revision, whether human listeners implement revision in the same form remains an open question.

\section{Conclusion}

We introduced a computational framework for studying conversational grounding as a preserve-or-revise problem. Across revision assumptions, a clear pattern emerged: successful grounding depends less on reacting to local mismatch than on selective revision under uncertainty. Counter to a common computational intuition, mismatch is associated with reduced revision, whereas uncertainty permits accumulated understanding to change. These findings suggest that conversational grounding is better understood as selective belief modulation than purely reactive repair, broadly consistent with conceptual pact theory. The framework provides a computational instrument for studying how understanding evolves through interaction.

\section{Limitations}

\textbf{Single dataset.} All experiments are conducted on PhotoChat, one of the few large-scale grounded dialogue corpora in which common ground must be constructed through conversation before image disclosure. While additional datasets would strengthen generalizability, few alternatives provide the combination of turn-level dialogue, hidden visual context, and progressive referential grounding required to evaluate preserve-or-revise dynamics. Because PhotoChat interactions are cooperative and goal-oriented, coherent grounding is typically maintained and strong repair events are relatively infrequent, potentially favoring persistence-based strategies. Whether the separation between local mismatch and successful revision generalizes to less stable communicative settings therefore remains open. As a next step, we plan to evaluate these revision policies on I-CONECT dataset \cite{dodge2024internet}, a caregiver-guided reminiscence dialogue corpus in which older adults, including individuals with mild cognitive impairment, discuss shared photographs. Compared with PhotoChat, this setting introduces less stable grounding and greater conversational scaffolding, providing a natural stress test for preserve-or-revise dynamics under cognitive and communicative pressure.

\textbf{Interpretability versus expressivity.}
Our design intentionally prioritizes interpretability over capacity, making preserve-or-revise dynamics explicit rather than maximizing performance. Whether richer architectures can retain this transparency while capturing finer-grained grounding remains an open question.

\bibliography{custom}

\appendix
\section{Appendix A: Implementation Details}
\label{sec:appendix}

\subsection{Training Protocol}

All grounding regimes share an identical frozen CLIP 
encoder (ViT-B/32) and optimization protocol, trained 
end-to-end with the following settings:

\begin{itemize}
    \item Optimizer: AdamW
    \item Learning rate: $5 \times 10^{-4}$
    \item Batch size: 16 dialogues
    \item Epochs: 12
    \item Temperature: $\tau = 0.07$
    \item Random seeds: $\{0, 1, 2\}$
\end{itemize}

The CLIP backbone remains frozen throughout. Results 
are reported as mean $\pm$ std over three seeds. The 
train--test split is fixed independently of the 
training seed (split seed $= 42$).

\subsection{Loss Hyperparameters}

All dynamic regimes (\textsc{fixed}, \textsc{dcp-b}, 
\textsc{multi-c}) share the following loss weights:

\begin{itemize}
    \item $\alpha$ (preserve): $0.1$
    \item $\beta$ (revise): $0.1$
    \item $\gamma_{\mathrm{rate}}$ (sparsity): $10.0$
    \item Target revision rate $r$: $0.15$
    \item DCP sharpness initialization $a$: $2.0$
\end{itemize}

\textsc{pure-corr} sets $\alpha = \beta = 0$, 
disabling all grounding dynamics losses.

\subsection{Data Split}

PhotoChat contains 21 training shards. We construct 
a deterministic split at the shard level: 19 shards 
for training, 2 for testing (split seed $= 42$). 
The split is created once and shared across all 
training seeds and regime variants.

\subsection{Reproducibility}

All six regime variants are trained in two parallel 
waves across three GPUs per wave. Wave 1 trains 
\textsc{fixed}, \textsc{dcp-b}, and \textsc{pure-corr} 
in parallel; Wave 2 trains \textsc{multi-c} and two 
ablated variants. Evaluation runs for all six trained 
models and two untrained baselines (Random, 
\textsc{clip-only}) are executed sequentially per 
seed. Per-seed results are aggregated into mean 
$\pm$ std tables using a fixed aggregation script. 
The full pipeline is idempotent: interrupted runs 
resume from the last completed checkpoint.

\section{Appendix B: Modality Ablation}
\label{sec:appendixb}

Table~\ref{tab:ablation} reports inference-time 
perturbation results for \textsc{multi-c}. Each 
modality is replaced independently with zero 
vectors, random Gaussian vectors, or empirical 
mean embeddings computed over the test set. No 
retraining is performed.

\begin{table}[h]
\centering
\scriptsize
\setlength{\tabcolsep}{4pt}
\begin{tabular}{lcccc}
\toprule
\textbf{Condition}
& \textbf{R@1}
& \textbf{R@5}
& \textbf{R@10}
& \textbf{MRR} \\
\midrule
Full
& 0.112 $\pm$ .004
& 0.303 $\pm$ .011
& 0.418 $\pm$ .009
& 0.212 $\pm$ .002 \\
\midrule
Img-Zero   & 0.000 $\pm$ .000 & 0.000 $\pm$ .000 & 0.000 $\pm$ .000 & 0.002 $\pm$ .000 \\
Img-Random & 0.001 $\pm$ .000 & 0.004 $\pm$ .003 & 0.010 $\pm$ .004 & 0.008 $\pm$ .001 \\
Img-Mean   & 0.000 $\pm$ .000 & 0.000 $\pm$ .000 & 0.000 $\pm$ .000 & 0.002 $\pm$ .000 \\
\midrule
Txt-Zero   & 0.000 $\pm$ .000 & 0.000 $\pm$ .000 & 0.000 $\pm$ .000 & 0.002 $\pm$ .000 \\
Txt-Random & 0.003 $\pm$ .002 & 0.006 $\pm$ .002 & 0.013 $\pm$ .001 & 0.010 $\pm$ .001 \\
Txt-Mean   & 0.001 $\pm$ .000 & 0.006 $\pm$ .000 & 0.011 $\pm$ .001 & 0.008 $\pm$ .000 \\
\bottomrule
\end{tabular}
\caption{
Inference-time modality ablation for \textsc{multi-c}
(mean $\pm$ std, three seeds). Replacing either
modality collapses retrieval to near-random,
confirming grounding depends jointly on both
modalities.
}
\label{tab:ablation}
\end{table}

\end{document}